\def\year{2021}\relax
\pdfoutput=1
\documentclass[letterpaper]{article} 
\usepackage{aaai21}  
\usepackage{times}  
\usepackage{helvet} 
\usepackage{courier}  
\usepackage[hyphens]{url}  
\usepackage{graphicx} 
\usepackage{natbib}  
\usepackage{caption} 
\usepackage{multirow} 
\usepackage[table]{xcolor} 

\usepackage[switch]{lineno}  

\title{UBAR: Towards Fully End-to-End Task-Oriented Dialog System with GPT-2}
\author {
        Yunyi Yang,
        Yunhao Li, 
        Xiaojun Quan\thanks{Xiaojun Quan is the corresponding author.} 
        \\
}
\affiliations {
    Sun Yat-sen University \\
    \{yangyy37, liyh355\}@mail2.sysu.edu.cn, quanxj3@mail.sysu.edu.cn
}

\begin{document}
\maketitle

\begin{abstract}

This paper presents our task-oriented dialog system UBAR
which models task-oriented dialogs on a dialog session level.
Specifically, UBAR is acquired by fine-tuning the large pre-trained unidirectional language model GPT-2 on the sequence of the entire dialog session which is composed of user utterance, belief state, database result, system act, and system response of every dialog turn.
Additionally, UBAR is evaluated in a more realistic setting, where its dialog context has access to user utterances and all content it generated such as belief states, system acts, and system responses.
Experimental results on the MultiWOZ datasets show that UBAR achieves state-of-the-art performances in multiple settings, improving the combined score of response generation, policy optimization, and end-to-end modeling by 4.7, 3.5, and 9.4 points respectively.
Thorough analyses demonstrate that the session-level training sequence formulation and the generated dialog context are essential for UBAR to operate as a fully end-to-end task-oriented dialog system in real life.
We also examine the transfer ability of UBAR to new domains with limited data and provide visualization and a case study to illustrate the advantages of UBAR in modeling on a dialog session level.\footnote{Code and technical appendix available at \url{https://github.com/TonyNemo/UBAR-MultiWOZ}}
\end{abstract}

\section{Introduction}
Task-oriented dialog (TOD) systems aim to assist users with various tasks such as hotel reservations and ticket booking through natural language conversations.
Recent years have seen a rapid growth of interest in developing data-driven approaches for this task from both the research community and industry \cite{zhang2020recent}.
The presence of wide range of domains requires TOD systems to have better transfer ability while remaining practical in real conversations.

\begin{figure}[]
    \centering
    \includegraphics[width=0.98\linewidth]{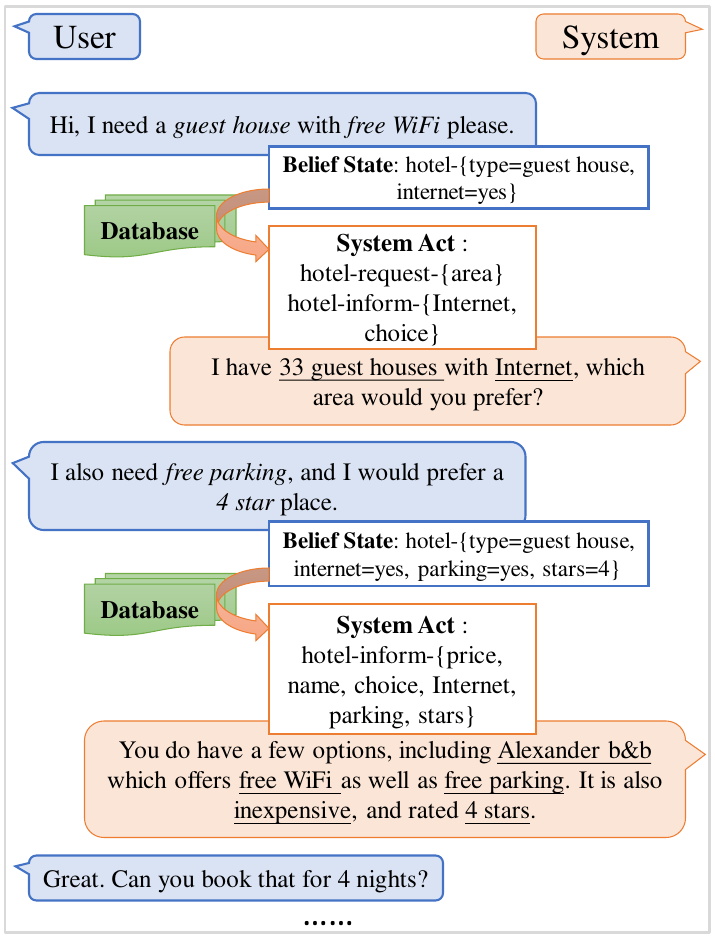}
    \caption{An example of the first two-turn interactions between a user and a TOD system.}
    \label{fig:dialog_example}
\end{figure}

The functions of a task-oriented dialog system can be understood by introducing the traditional pipeline approach which consists of several consecutive modules.
As shown in Figure \ref{fig:dialog_example}, a dialog state tracker (DST) is equipped  to estimate the belief state from the user utterance. The belief state can be used to query a task-related database (DB) for results such as the number of entities that match the user's goal. Then, a dialog policy learning module is applied to determine the next system act, followed by a natural language generation (NLG) module that maps the system act to a natural language response.
These modules are often modeled and evaluated separately.
The apparent drawback of the pipeline approach is that error propagation from the cascaded components can be detrimental to the subsequent subtasks \cite{liu2018end}.

From a big picture perspective, the methodology for task-oriented dialog systems is gradually progressing from pipeline approaches to end-to-end modeling.
Recently, some work attempts to incorporate the intermediate supervision, i.e., the belief state and system act, and train systems in joint learning settings \cite{chen2019semantically,wang2020multi}. They jointly generate system act and response, yet still using ground truth belief state.
Some methods have come close to modeling TOD in an end-to-end manner, but they tend to use different decoders for each component.
For example, \citet{lei2018sequicity} and \citet{liang2020moss} use a seq2seq model to generate belief spans and responses.
\citet{zhang2020task} propose multiple decoders to generate belief spans, act spans, and responses.

On the other hand, the large pre-trained language model GPT-2 \cite{radford2019language} is shown to be capable of modeling the dialog pipeline in a unified way.
Initially, \citet{ham2020end} propose to train a unified language model for task-oriented dialogs with a single sequence in format of dialog history (all previous user utterances and responses), user utterance, belief state, system act, response of the current turn, and evaluate for DST and policy optimization.
SimpleTOD \cite{hosseini-asl2020a} and SOLOIST \cite{peng2020soloist} further generalize this idea to an end-to-end setting where the belief states are also generated instead of using ground truth values. They also incorporate database results into the training process.
In spite of the promising results from leveraging pre-trained language models like GPT-2 for end-to-end TOD systems, these methods do not fully explore the process of training and evaluating towards a real-life task-oriented dialog setting.
Specifically, these GPT-2-based TOD systems are trained and evaluated on a dialog turn level instead of the dialog session level, which has several limitations.
First, the dialog history of these methods only consists of user utterances and system responses but leaves out the intermediate information such as belief states and system acts of the previous turns.
These intermediate information could be a helpful reference for the generation of the current turn.
Second, they use the ground truth responses from annotations in the dialog history, which makes the generation of a dialog turn independent of other turns in a dialog session.
Third, the assumption of having access to the ground truth system responses is invalid in real conversations.

To address the aforementioned limitations and advance towards a fully end-to-end TOD system, we propose UBAR to model task-oriented dialogs on a dialog session level.
We fine-tune GPT-2 on the sequence of the entire dialog session consisting of user utterance, belief state, database result, system act, and system response of every dialog turn.
Such training data formation resembles the workflow of a real-life task-oriented dialog session, which allows UBAR to learn task completion and language generation over the course of a dialog session.
UBAR is able to condition on the previous belief states and system acts in the dialog context, making the process of inference and planning easier for the current turn.
Since in real conversations, a TOD system should be able to access the belief states it predicted and the system acts and responses it generated throughout the entire dialog session.
We further propose to evaluate UBAR with the dialog context of generated content instead of the ground truth.
This encourages UBAR to adaptively supplement and make amends in response to the current user utterance in order to stay consistent and coherent during the entire session, and ultimately contribute to the task completion goal.

We conduct experiments on the MultiWOZ datasets \cite{budzianowski2018multiwoz,eric2019multiwoz} in multiple settings including response generation, policy optimization, end-to-end modeling and dialog state tracking, and compare UBAR with its GPT-2-based predecessors and other strong baselines.
UBAR achieves state-of-the-art performances in all compared settings.
We perform thorough analysis to show that the session-level training sequence formulation and all-generated dialog context are essential for UBAR to operate as a fully end-to-end TOD system in real life.
We also examine the transfer ability of UBAR to new domains given limited data, and provide visualization and case study to illustrate the advantages of modeling task-oriented dialogs on a session level.

\begin{figure*}[ht]
    \centering
    \includegraphics[width=\textwidth]{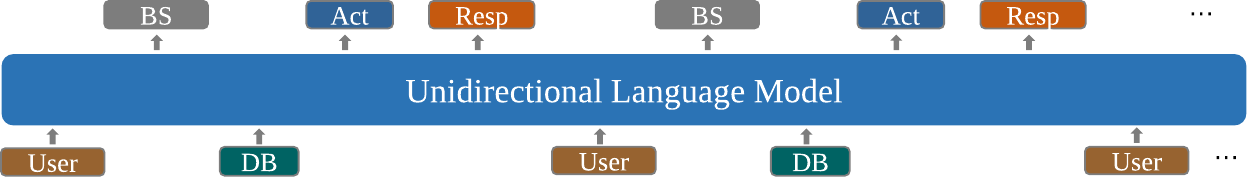}
    \caption{An overview of UBAR.
    }
    \label{fig:model}
\end{figure*}

\section{Related Work}
\subsection{Towards End-to-End Task-Oriented Dialog}
With the emergence of large-scale multi-domain TOD datasets \cite{budzianowski2018multiwoz,shah2018bootstrapping,peskov2019multi}, the methodology for task-oriented dialog systems can be roughly seen to gradually progress from classification and modularized modeling to generation and end-to-end modeling over the recent years.
Early methods for DST are commonly formulated as a classification task, where the dialog state representation maintains a distribution over all possible states for each slot \cite{henderson2013deep,henderson2014word,zhang2019neural}.
To generalize to tracking unknown slot values and multi-domain settings, generative methods are proposed to extract slot values for DST \cite{zhong2018global,xu2018an,wu2019transferable}.
Similarly for dialog policy learning, system acts are originally encoded in vector representations such one-hot vectors and used for response generation \cite{chen2019semantically,zhao2019rethinking,wen2017latent}.
Then, they are jointly trained and generated with system responses \cite{wang2020multi,zhang2020task}.
For end-to-end modeling, \citet{lei2018sequicity} propose a two-stage CopyNet \cite{gu2016incorporating} that generates belief spans and system response jointly via a single seq2seq architecture.
\citet{zhang2020task} propose a domain-aware multi-decoder model that uses separate decoders to generate belief spans, act spans and responses.
Recently, pre-trained language model like GPT-2 is also leveraged for end-to-end modeling in a unified way \cite{peng2020soloist,hosseini-asl2020a}.
Besides, end-to-end TOD systems that directly operate on dialog history and interact with knowledge base without any intermediate supervision \cite{eric2017key,madotto2018mem2seq,wu2019global} also receive growing attention, but are not within the scope of our discussion.

\subsection{Pre-trained Language Models for Dialog Systems}
Large pre-trained language models have shown superior performance on a wide range of NLP tasks \cite{peters2018deep,devlin2019bert}, and GPT-2 \cite{radford2019language} is especially good at language generation tasks.
Some work extends GPT-2 \cite{radford2019language} to generate responses in chit-chat dialog \cite{zhang2020dialogpt,wu2020a}.
In task-oriented dialog domain, \citet{budzianowski2019hello} first point out the possibility of fine-tuning all necessary information in simple text on GPT-2 which inspires a line of improved and simplified design of task-oriented dialog systems.
SC-GPT \cite{peng2020few} is a pre-trained model that converts ground-truth system acts into responses.
\citet{ham2020end} fine-tune GPT-2 in a similar fashion for DST and policy optimization, but employ heuristic rules to handle different database query results.
SimpleTOD \cite{hosseini-asl2020a} incorporates the database results into the training process and is evaluated for end-to-end modeling where belief state and system act are generated.
SOLOIST \cite{peng2020soloist} follows a pre-train and fine-tune paradigm where it first undergoes pre-training on a large number of out-of-domain dialog turns, then fine-tune on the data of new domains. It does not require the annotation of system acts.
This work follows its GPT-2-based predecessors and progresses for a fully end-to-end TOD system by operating in terms of a whole dialog session instead of a dialog turn during training and evaluating.

\section{Method}
In this section, we describe how UBAR models on a dialog session level and how we prepare the dialog data to be trained in sequence. Figure \ref{fig:model} is an overview of UBAR.

\subsection{Modeling on a Dialog Session Level}
The workflow of a TOD system interacting with a user naturally produces a sequence as it reads user utterances, tracks dialog states and generates acts and responses over the turns of a dialog session.

Given a dialog session composed of multiple turns, we show how UBAR models the process of a task-oriented dialog session. In the first turn $t=0$, the user inputs user utterance $U_0$, UBAR generates a belief state $B_0$ based on $U_0$. This belief state is used to query a database to retrieve the matched number of entities that satisfy the constraint imposed by the belief state, which is the database search result $D_0$.
Conditioned on $\{U_0, B_0, D_0\}$, UBAR then generates system act $A_0$ and the delexicalized response $R_0$, completing the interaction of the first turn.
As the dialog proceeds to turn $t$, UBAR generates $B_t$, $A_t$ and $R_t$ based on context of user utterances and all previous generated outputs $\{U_0, B_0, D_0, A_0, R_0, ... , U_{t-1}, B_{t-1}, D_{t-1}, A_{t-1}, R_{t-1}, U_{t}\}$, eventually completing the entire dialog session.
Therefore, a single training sequence for a dialog session with $T$ turns can be formulated as $\{U_0, B_0, D_0, A_0, R_0, ... ,$ $U_T, B_T, D_T, A_T, R_T\}$.

Note that UBAR is different from previous methods that are trained and evaluated in turn-level sequences where they are based on dialog history $\{U_0,R_0,...,U_{t-1},R_{t-1},U_{t}\}$ to generate the response in turn $t$ and their responses in the dialog history are ground truth. While UBAR incorporates the intermediate information $B$, $D$ and $A$ in the context.

\subsection{Domain-Adaptive Pre-processing}
\begin{figure}[ht]
    \centering
    \includegraphics[width=\linewidth]{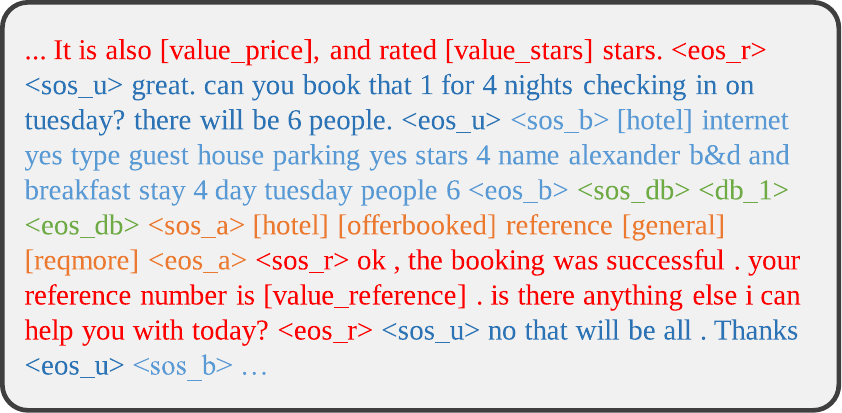}
    \caption{Example of different (colored) components in a dialog session concatenated into a single training sequence.}
    \label{fig:training_data}
\end{figure}

We pre-process the dialog data into the form of simple text by delexicalizing the responses and converting intermediate dialog information to spans. All sequences are surrounded by special $<$\textit{sos\_?}$>$ and $<$\textit{eos\_?}$>$\footnote{The \textit{?} represents a component from \{u, b, db, a, r\}.} tokens indicating the beginning and end of a component. Figure \ref{fig:training_data} is an exemplar pre-processed training sequence.

\noindent\textbf{Delexicalization}
It is important to generate delexicalized responses for task-oriented generation, which allows the model to learn value-independent parameters \cite{wen2015semantically}. Delexicalization replaces specific slot values by their corresponding placeholders, which can be filled according to database search results afterwards.  We adopt a domain-adaptive delexicalization scheme \cite{zhang2020task} to decouple the domain and slot name of placeholders. For example, a hotel name in the generated response is just \textit{$<$value\_name$>$} instead of \textit{{$<$hotel-value\_name$>$}}.

\noindent\textbf{Belief State and System Act Spans}
Belief states are originally represented in \textit{domain-slot, value} pairs.
We decouple domain and slot names to allow generalization cross different domains that share the same ontology.
A schematic example of belief state span with two mentioned domains would be \{\textit{[domain1] slot value slot value [domain2] slot value}\}.
Each domain can follow several slot value pairs specifying user's goal.

For database search results, we use special tokens to indicate the number of matched entities under the constraints of the belief state of the current turn.

System acts are originally represented as \textit{domain-act, slot} pairs which aim to inform or request information about the slot of a certain domain.
We also decouple the domain and act for system act span: \{\textit{[domain] [inform] slot ...  [request] slot ...}\}.
The decoupling of domains allows dialog ontology as well as expressions to be learned across relevant domains.
The domains, acts and slot values are all bracketed as additional special tokens so that they can be learned specifically.

\subsection{Architecture and Training Objective}
GPT-2 \cite{radford2019language} is a powerful pre-trained unidirectional language model. It is a large Transformer decoder \cite{vaswani2017attention}
that is trained on large corpora of web text and can generate realistic and coherent natural language.
By fine-tuning GPT-2 on session-level task-oriented dialog data, UBAR learns to ground generation with ontology knowledge and decision making ability.

The training objective for UBAR is the language modeling objective \cite{bengio2003a}, which maximizes the probability of next word prediction:
$L = \sum_i log P(w_i|w_{<i})$.
UBAR does not require additional training objectives such as next-utterance classification.

\begin{table*}[t]
    \centering
    \small
    \begin{tabular}{l|c c | c c c c}
        \hline
        Model & Belief State & System Act & Inform & Success & BLEU & Combined\\
        \hline
        HDSA & oracle & oracle & 87.9 & 78.0 & \textbf{30.4} & 113.4 \\
        DAMD & oracle & oracle & 95.4 & 87.2 & 27.3 & 118.5 \\
        SimpleTOD & oracle & oracle & 92.3 & 85.8 & 18.67 & 107.7 \\
        UBAR (ours) & oracle & oracle & \textbf{96.9} & \textbf{92.2} & 28.6 & \textbf{123.2} \\
        \hline
        SFN+RL & oracle & generated & 82.7 & 72.1 & 16.3 & 93.7 \\
        HDSA & oracle & generated & 82.9 & 68.9 & \textbf{23.6} & 99.5 \\
        ARDM & oracle & - & 87.4 & 72.8 & 20.6 & 100.7 \\
        DAMD & oracle & generated & 89.2 & 77.9 & 18.6 & 102.2 \\
        SimpleTOD & oracle & generated & 88.9 & 67.1 & 16.9 & 94.9 \\
        SOLOIST & oracle & - & 89.6 & 79.3 & 18.0 & 102.5 \\
        UBAR (ours) & oracle & generated & \textbf{94.0} & \textbf{83.6} & 17.2 & \textbf{106.0} \\
        \hline
        SFN+RL & generated & generated & 73.8 & 58.6 & 16.9 & 83.0 \\
        DAMD & generated & generated & 76.3 & 60.4 & 16.6 & 85.0 \\
        SimpleTOD & generated & generated & 84.4 & 70.1 & 15.0 & 92.3 \\
        SOLOIST & generated & - & 85.5 & 72.9 & 16.5 & 95.7 \\
        UBAR (ours) & generated & generated & \textbf{95.4} & \textbf{80.7} & \textbf{17.0} & \textbf{105.1} \\
        \hline
    \end{tabular}
    \caption{Comparison of generation results on MultiWOZ 2.0. The oracle/generated denotes either using ground truth or generated intermediate information. The results are grouped according to how belief state and system act are modeled.}
    \label{tab:overall_e2e}
\end{table*}

\section{Experiments}
\subsection{Dataset and Evaluation Metrics}
MultiWOZ 2.0 \cite{budzianowski2018multiwoz} is a large-scale human-to-human multi-domain task-oriented dialog dataset consisting of 8438 dialogues spanning over seven domains (\textit{attraction, hospital, police, hotel, restaurant, taxi, train}). It provides additional validation set and test set each of 1000 dialogues, excluding hospital and police. Each dialog session contains 1 to 3 domains and multiple domains might be mentioned in a single turn (more dataset details in appendix).
MultiWOZ 2.1 \cite{eric2019multiwoz} is an improved version of MultiWOZ 2.0 by fixing some noisy state annotations.
We conduct experiments and analyses on the 2.0 version and also report results on the 2.1 version.

We follow the automatic evaluation metrics to evaluate task completion and response quality: \textbf{Inform} measures whether a system has provided a correct entity, \textbf{Success} measures whether it has answered all the requested information, and \textbf{BLEU} \cite{papineni2002bleu} is used to measure the fluency of the generated responses \cite{budzianowski2018multiwoz}. A combined score: $(\textbf{Inform} + \textbf{Success}) \times 0.5 + \textbf{BLEU}$ is also reported as an overall quality measure suggested in \citet{mehri2019structured}.
We also use the joint goal accuracy to evaluate dialog state tracking (DST).
\subsection{Implementation Details}
We implement UBAR with HuggingFace's Transformers \cite{wolf2019huggingface} and DistilGPT2 \cite{sanh2019distilbert}, a distilled version of GPT-2.
The model is trained on session-level sequences with a max sequence length of 1024. Sequences that exceed 1024 tokens are pre-truncated.
We use the AdamW optimizer and standard greedy decoding method with temperature of 0.7.
We select the best performing model on validation set through hyperparameters search of learning rate and batch size, then evaluate on test set to get the final results.
We also report the performances of UBAR on validation set in technical appendix.
Code and models are included in the supplement and will be released.

\subsection{Baselines}
We compare UBAR with SimpleTOD \cite{hosseini-asl2020a} and SOLOIST \cite{peng2020soloist}, the GPT-2-based methods that are trained on turn-level data without generated belief state and system act in dialog history \cite{hosseini-asl2020a,peng2020soloist}, and other several competitive methods HDSA \cite{chen2019semantically}, SFN+RL \cite{mehri2019structured}, ARDM \cite{wu2019alternating}, and DAMD \cite{zhang2020task}.
UBAR is evaluated and compared in three context-to-response settings: response generation based on ground truth belief state and system act, policy optimization to generate system act and response based on ground truth belief state, and end-to-end modeling to generate belief state, system act and response.
Experiments with ground truth belief state use ground truth database search result.
All content UBAR generated during a dialog session will remain in the dialog context for the generation the current turn.

Since the proposed UBAR can generate belief state throughout the entire dialog session, we compare the performance of UBAR on dialog state tracking with GPT-2-based model SimpleTOD \cite{hosseini-asl2020a} and other state-of-the-art methods such as TRADE \cite{wu2019transferable}, DSTQA \cite{zhou2019multi}, DST-Picklist \cite{zhang2019find}, SST \cite{chen2020schema}.
As DST requires extracting slot values from non-delexicalized responses, we train a new DST-UBAR using non-delexicalized responses for DST evaluation.

\subsection{Overall Results}
\noindent \textbf{Response Generation with Ground truth Belief State and System Act}
The first group in Table \ref{tab:overall_e2e} shows the results of response generation based on the ground truth belief state and system act. UBAR applies the same domain-adaptive delexicalization and domain-aware belief, act spans as previous state-of-the-art DAMD, yet outperforms all compared methods in response generation in terms of inform rate, success rate and combined score, including DAMD.
The BLEU score is slightly lower but inform rate and success rate are much higher than HDSA, which indicates UBAR is more grounded in task completion than language surface.

\noindent \textbf{Policy Optimization with Ground truth Belief State}
In the setting of policy optimization, the context of UBAR consists of ground truth belief states and database results and generated act and responses.
As shown in the second group of Table \ref{tab:overall_e2e}, UBAR achieves the best performance in terms of inform rate, success rate and combined score, improving the previous state-of-the-art SOLOIST by a large margin (3.5 points on the combined score).
Note that SOLOIST is first pre-trained on large-scale task-oriented dialog data and then fine-tuned on MultiWOZ, and they did not leverage the system act.
UBAR achieves higher inform and success rate than SOLOIST without additional pre-training data.
These results show that UBAR can effectively learn dialog policy and response generation.

\noindent \textbf{End-to-end Modeling}
The third group in Table \ref{tab:overall_e2e} shows results in end-to-end modeling setting, where UBAR has to generate belief state, query database result with the generated belief state, and then generates act and response.
UBAR achieves the state-of-the-art performance on all metrics and lifts almost 10 points on the combined score.
Like SimpleTOD and SOLOIST, UBAR has a very simple architecture and is trained on sequences with language modeling objective.
Unlike SimpleTOD and SOLOIST, UBAR uses all generated content in dialog context instead of ground truth responses.
UBAR demonstrates incredible ability in modeling a complete task-oriented dialog session in an arguably fully end-to-end fashion, much closer to a task-oriented conversation in real life.

\noindent \textbf{Results on MultiWOZ 2.1}
We also report the performance of UBAR in the three settings on MultiWOZ 2.1 for future comparison. As shown in Table \ref{tab:MW2.1}, the results are consistent with that on MultiWOZ 2.0.

\begin{table}[t]

    \small
    \centering
    \begin{tabular*}{\columnwidth}{c c | c c c c}
        \hline
        Belief &  Act & Inf. & Succ. & BLEU & Comb. \\
        \hline
        oracle & oracle & 95.4 & 91.4 & 28.8 & 122.2\\
        oracle & generated & 92.7 & 81.0 & 16.7 & 103.6\\
        generated & generated & 95.7 & 81.8 & 16.5 & 105.7\\
        \hline
    \end{tabular*}
    \caption{UBAR in different settings on MultiWOZ 2.1.}
    \label{tab:MW2.1}
\end{table}

\begin{table}[t]
    \small
    \centering
    \begin{tabular}{l|c|c}

        \hline
        \multirow{2}{3em}{Model} & \multicolumn{2}{c}{Joint Accuracy (\%)} \\
        \cline{2-3}
        & MultiWOZ 2.0 & MultiWOZ 2.1 \\
        \hline
        TRADE & 48.62 & 45.60 \\
        DSTQA & 51.44 & 51.17 \\
        DST-Picklist & - & 53.3\\
        SST & - & 55.23 \\
        SimpleTOD & - & 55.72 \\
        DST-UBAR & \textbf{52.59} & \textbf{56.20}\\
        \hline
    \end{tabular}
    \caption{Comparison of Dialog state tracking (DST) on MultiWOZ 2.0 and 2.1.}
    \label{tab:DST}
\end{table}

\noindent \textbf{Dialog State Tracking}
As shown in Table \ref{tab:overall_e2e}, The small gap between the performances in policy optimization and end-to-end modeling suggests that UBAR is relatively good at generating belief states.
Even though UBAR is designated for response generation and end-to-end modeling, we evaluate the design of UBAR on dialog state tracking task by training a DST-UBAR with training data consisting of user utterances, belief states, database results, system acts and \textbf{non-delexicalized} responses of dialog sessions.
The belief states are generated by DST-UBAR based on dialog context of user utterances, generated belief states and ground truth system acts and non-delexicalized responses.
Table \ref{tab:DST} shows that DST-UBAR also outperforms the state-of-the-art methods on both versions of MultiWOZ.

\section{Analysis and Discussion}
In this section, we try to answer three questions:
(1) How much and what kind of dialog context does UBAR need for end-to-end modeling? (2) What are the advantages of the proposed training and evaluating on a dialog session level over turn level? (3) How well can UBAR transfer to unseen domains in end-to-end modeling?

\subsection{Dialog Context}
A large portion of the information of user's goal is stored in belief states.
Since UBAR incorporates belief states in the context, it can figure out the new belief states based on just the previous turn and user utterance of the current turn.
As shown in Table \ref{tab:context}, UBAR based on the previous turn underperforms UBAR based on all previous turns, but still outperforms other state-of-the-art methods.
Therefore, UBAR can operate properly with much shorter context length than turn-level methods that require full dialog history, which is more computationally efficient.
On the other hand, if UBAR is granted with ground truth belief states in the context, the results would increase slightly.
This is because the ground truth belief states in the context make generating belief states of the current turn easier.
However, if UBAR takes all ground truth content in the context including system acts and responses, the results actually drop quiet a lot.
This is somewhat unexpected yet understandable given the acts and responses in the context are not committed by UBAR, and could mislead UBAR to think that it already committed such acts and responses.
We will discuss more on this in the case study next section.
In a realistic setting, a TOD system can have access to the context it generated, but not any ground truth, which is why we report all-generated results for overall comparison.

\begin{table}[t]
    \small
    \centering
    \begin{tabular*}{\columnwidth}{c | c | c | c c c c}
        \hline
        \#Turns & Belief &  Act & Inf. & Succ. & BLEU & Comb. \\
        \hline
        All & GT & GT & 88.4 & 76.6 & \textbf{17.6} & 100.1\\
        All & GT & Gen & 95.4 & \textbf{82.3} & 17.2 & \textbf{106.1} \\
        \rowcolor{lightgray} All & Gen & Gen & \textbf{95.4} & 80.7 & 17.0 & 105.1 \\
        \hline
        Prev & GT & GT & 87.2 & 75.3 & \textbf{16.8} & 98.0 \\
        Prev & GT & Gen & 92.7 & \textbf{79.0} & 16.6 & \textbf{102.5} \\
        \rowcolor{lightgray} Prev & Gen & Gen & \textbf{92.7} & 77.7 & 16.4 & 101.6 \\
        \hline
    \end{tabular*}
    \caption{Results of UBAR evaluated with different kinds of dialog context in end-to-end setting. \emph{\#Turns} denotes the number of previous turns in context, \emph{All} means all previous turn, \emph{Prev} means just the last turn.
    \emph{GT} or \emph{Gen} denotes if the belief states and system acts in the context are ground truth or generated. We provide a more comprehensive evaluation of UBAR with different kinds of dialog context in multiple settings in the technical appendix.}
    \label{tab:context}
\end{table}

\subsection{Session-Level vs. Turn-Level}
The main difference between UBAR and other GPT-2-based models is that UBAR is trained on session-level sequences with intermediate information such as belief states and system acts in the context, while others are trained on turn-level sequences with only dialog history of user utterances and system responses.
To study the effect of session-level training and the incorporation of belief states and system acts in the context via ablation, we implement a model URUR trained on turn-level sequences.
We evaluate URUR in the end-to-end modeling setting where every turn it makes generation conditioned on previous user utterances and system responses.
As shown in Table \ref{tab:URUR}, the turn-level URUR underperforms UBAR.
Specifically, URUR with ground truth responses in dialog history has comparable performance with SimpleTOD and SOLOIST.
What's more, URUR with generated responses in history shows significant improvement over ground truth responses, which suggests that SimpleTOD and SOLOIST miss out a convenient boost evaluating on a turn-level.
They could achieve better performances by simply using generated responses in their dialog context.

On the other hand, we constrain UBAR to generate based on the context consisted of only user utterances and responses or only belief states and system acts. With B\&A outperforming U\&R as well as URUR, we confirm that the belief states and system acts are more important than user utterances and responses in dialog context and that it is more difficult for models to infer belief states and system acts from the dialog history of every turn.

\begin{table}[t]
    \small
    \centering
    \begin{tabular}{l | c | c c c c}
        \hline
        Model & Context & Inf. & Succ. & BLEU & Comb. \\
        \hline
        URUR & GT & 82.6 & 73.1 & \textbf{17.0} & 94.8\\
        URUR & Gen & \textbf{91.2} & \textbf{79.5} & 16.5 & \textbf{101.8}\\
        \hline
        UBAR & U\&R & 92.5 & 70.8 & 14.3 & 95.9\\
        UBAR & B\&A & \textbf{94.1} & \textbf{77.1} & \textbf{16.3} & \textbf{101.9}\\
        \hline
    \end{tabular}
    \caption{Results in the end-to-end setting. URUR is trained in turn-level. \emph{GT} or \emph{Gen} means it uses ground truth or generated responses in its context. \emph{U\&R} denotes the context of UBAR only consists of user utterances and generated responses. \emph{B\&A} denotes the context of UBAR only consists of belief states, database results and system acts.}
    \label{tab:URUR}
\end{table}

\begin{table*}[t]
    \small
    \centering
    \begin{tabular}{c|c | c |c | c | c}
        \hline
        \textbf{Evaluation on 4 Domains} & Except Hotel & Except Train & Except Attraction & Except Restaurant & Except Taxi \\
        \hline
        Base Model trained in-domain & 99.03 & 99.40 & 99.68 & \textbf{101.87} & 95.30 \\
        \hline
        Few-shot BM on new domain & 84.88 & 84.66 & 96.96 & \textbf{100.79} & 88.14 \\
        \hline\hline
        \textbf{Evaluation on New Domain} & Hotel & Train & Attraction & Restaurant & Taxi \\
        \hline
        Zero-shot BM & 58.40 & \textbf{64.19} & 49.96 & 40.55 & 59.03 \\
        \hline
        Few-shot BM on new domain & 78.48 &  70.52 &  \textbf{79.51} & 74.19 & 74.02 \\
        \hline
        UBAR on all domains & 103.06 & \textbf{106.40} & 102.24 & 104.44 & 103.20 \\
        \hline
    \end{tabular}
    \caption{Results of domain transfer. The first row is the base model trained on the four domains and evaluated in-domain. The second row is the results of the base model fine-tuned with 100 new domain examples on the four domains. The last three rows are evaluations on the new domains with zero-shot or few-shot BM or UBAR trained on full data, respectively.}
    \label{tab:transfer}
\end{table*}
\subsection{Domain Transfer}
The ontology are often shared across domains. For example, \textit{Hotel} and \textit{Restaurant} share the same requestable slots such as address, postcode, price range. Therefore, it is possible for UBAR to generalize to new domains.

To examine the transfer ability of UBAR to generalize to unseen domains, we run zero-shot and few-shot end-to-end modeling experiments by excluding one domain out of the five domains that are available in validation and test set, and training UBAR on other four domains.
As shown in Table \ref{tab:transfer}, after trained on 4 domains, the base model (BM) performs generally well in-domain.
In zero-shot setting, the performances vary across different domains.
The \textit{Train} domain achieves the highest combined score, while \textit{Restaurant} performs badly. This is because \textit{Train} domain has a high overlap in ontology with other domains, while \textit{Restaurant} has a unique food slot which is mentioned frequently.

In few-shot setting, the base model is fine-tuned with 100 dialog sessions from the held-out domain. The few-shot BM is evaluated on the held-out domain and achieves better performance than zero-shot BM, improving the combined score by 20 points in average, which demonstrates the transfer ability of UBAR.
However, we also see a drop in the performance of the few-shot BM evaluated on the original four domains, which indicates catastrophic forgetting to some extent.
What's more, the big gap between few-shot BM and UBAR trained on all domains underscores the data hungry nature of end-to-end task-oriented dialog modeling.

\begin{figure*}[ht]
    \centering
    \includegraphics[width=1\textwidth]{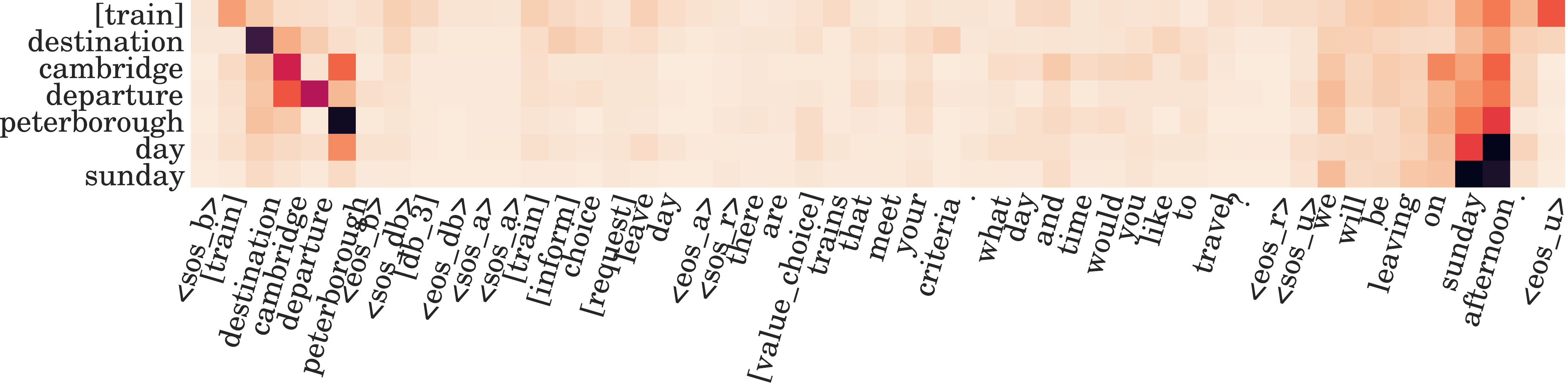}
    \caption{Visualization of attention weights of the generated belief states attending to the context. The X-axis is the belief states, response from the previous turn and the current user utterance. The Y-axis is the generated belief states.}
    \label{fig:visualization}
\end{figure*}

\begin{figure}[h!]
    \centering
    \includegraphics[width=\linewidth]{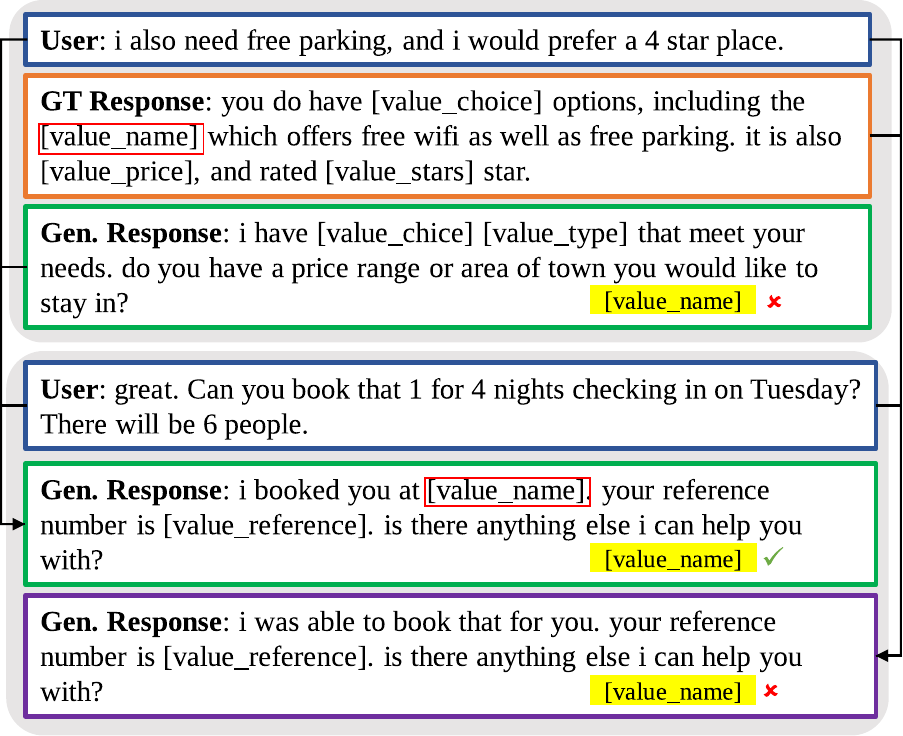}
    \caption{Two consecutive dialog turns in dialog session SNG0855 from MultiWOZ2.0.
    The green boxes and purple box indicate responses of UBAR based on generated context and response based on oracle context, respectively.
    The task related entities are highlighted in yellow.}
    \label{fig:case}
\end{figure}

\section{Visualization and Case Study}
In this section we provide an attention weights visualization to demonstrate how UBAR generates belief states based on user utterance and the belief states of the previous turn, and a case study to explain why UBAR can improve task completion by conditioning on the generated context. More visualizations and case studies are in the technical appendix.

Figure \ref{fig:visualization} visualizes the attention weights of the last layer of the transformer blocks in UBAR, demonstrating that UBAR appropriately attends to the belief state from the previous turn and copies those that remain consistent, which are the destination and departure of a train.
Then it attends to user utterance for an update with new specifications, which is the day of the train. It makes sense to keep track of the belief states and add in or make adjustment to them given incoming user utterance, instead of going through all previous user utterances and responses every turn.



Figure \ref{fig:case} shows a case where the user wants to book a 4-stars hotel with free parking.
In the first turn, UBAR did not provide the hotel entity, but asked for additional information about the hotel's price range or area.
However in the second turn, the user directly requests to book the unmentioned hotel.
If using the generated content, UBAR would provide a specific hotel entity in response to the user utterance in the second turn.
While if using the ground truth response which contains the hotel entity as the dialog context, UBAR would be mistaken that it had already provided such entity, failing to mention the important entity.

This explains why UBAR with generated context can outperform UBAR with ground truth context, as UBAR can adaptively supplement and make amends in response to the current user utterance in order to stay consistent and coherent throughout the entire session, ultimately improving task completion. 

\section{Conclusion}
In this paper, we attempt to approach end-to-end task-oriented dialog system in a more realistic setting.
The proposed UBAR is trained and evaluated on a dialog session level.
It generates belief states, system acts and responses based on the user utterances and all content it generated.
We conduct extensive experiments and analyses to demonstrate the superiority of the modeling on a dialog session level and the power of GPT-2.
We hope that UBAR can inspire more future work to model task-oriented dialog system on a session level.
\section*{Acknowledgments}
We thank the anonymous reviewers for their constructive reviews. This work was supported by the Fundamental Research Funds for the Central Universities (No.19lgpy220) and the Program for Guangdong Introducing Innovative and Entrepreneurial Teams (No.2017ZT07X355).

\bibliographystyle{aaai21}
\bibliography{references}
\end{document}